\documentclass[runningheads]{llncs}
\usepackage{graphicx}

\usepackage{hyperref}
\hypersetup{
    colorlinks=true,
    linkcolor=blue,
    filecolor=magenta,      
    urlcolor=cyan,
    pdfpagemode=FullScreen,
    }
    
\usepackage{times}
\usepackage{epsfig}
\usepackage{amsmath}
\usepackage{mathtools}
\usepackage[export]{adjustbox} 
\usepackage{booktabs} 
\usepackage{amsfonts}
\usepackage{subfigure}
\usepackage{multirow}
\usepackage{cleveref}
\usepackage{xcolor}
\usepackage{xspace}
\usepackage{dsfont}
\usepackage{eucal}
\usepackage{siunitx}
\usepackage{cleveref}
\crefformat{footnote}{#2\footnotemark[#1]#3}

\usepackage{algorithm}
\usepackage{algpseudocode}
\usepackage{vwcol}  

\usepackage{caption}
\usepackage{varwidth}

\DeclareCaptionLabelFormat{andtable}{#1~#2  \&  \tablename~\thetable}

\newcommand{\argmin}[1]{\underset{#1}{\operatorname{arg}\,\operatorname{min}}\;}

\usepackage{cleveref}
\crefrangelabelformat{figure}%
      {#3#1#4--#5\crefstripprefix{#1}{#2}#6}
\crefrangelabelformat{subfigure}%
      {#3#1#4#5-(\crefstripprefix{#1}{#2}#6}
\Crefname{figure}{Fig.}{Figs.}

\makeatletter
\def\@xfootnote[#1]{%
  \protected@xdef\@thefnmark{#1}%
  \@footnotemark\@footnotetext}
\makeatother

\begin{document}
\title{Contrastive Supervised Distillation for \\ Continual Representation Learning }

\renewcommand{\thefootnote}{\fnsymbol{footnote}}
\author{Tommaso Barletti$^{*}$\orcidID{0000-0001-7460-4710},
Niccol\'o Biondi$^{*\dagger}$\orcidID{0000-0003-1153-1651}, \\ 
Federico Pernici\orcidID{0000-0001-7036-6655},  
Matteo Bruni\orcidID{0000-0003-2017-1061}, \\ and 
Alberto Del Bimbo\orcidID{0000-0002-1052-8322}
}
\authorrunning{T. Barletti et al.}
\institute{Media Integration and Communication Center (MICC), Dipartimento di Ingegneria dell'Informazione, Università degli Studi di Firenze \\
\email{name.surname@unifi.it}
}
\vspace{-30pt}
\maketitle             

\footnotetext[1]{Tommaso Barletti and Niccol\'o Biondi contributed equally.}
\footnotetext[4]{Corresponding Author.}

\vspace{-20pt}
\renewcommand{\thefootnote}{\arabic{footnote}}
\begin{abstract}
In this paper, we propose a novel training procedure for the continual representation learning problem in which a neural network model is sequentially learned  to alleviate catastrophic forgetting in visual search tasks.
Our method, called  \textit{Contrastive Supervised Distillation} (CSD), reduces feature forgetting while learning discriminative features. This is achieved by leveraging labels information in a distillation setting in which the student model is contrastively learned from the teacher model.
Extensive experiments show that CSD performs favorably in mitigating catastrophic forgetting  by outperforming current state-of-the-art methods.
Our results also provide further evidence that  feature forgetting evaluated in visual retrieval tasks is not as catastrophic as in classification tasks.
Code at: \href{https://github.com/NiccoBiondi/ContrastiveSupervisedDistillation}{https://github.com/NiccoBiondi/ContrastiveSupervisedDistillation}.

\end{abstract}

\section{Introduction}
Deep Convolutional Neural Networks (DCNNs) have significantly advanced the field of visual search or visual retrieval by learning powerful feature representations from data \cite{wan2014deep,azizpour2014cnn,yosinski2014transferable}. 
Current methods predominantly  focus on learning feature representations from static datasets in which all the images are available during training \cite{chen2021deep,tolias2016particular,yue2015exploiting}. 
This operative condition is restrictive in real-world applications since new data are  constantly emerging and repeatedly training DCNN models on both old and new images is time-consuming. Static datasets, typically stored on private servers, are also increasingly problematic because of the societal impact associated with privacy and ethical issues of modern AI systems~\cite{price2019privacy,cossu2021sustainable}. 

These problems may be significantly reduced in incremental learning scenarios as the computation is distributed over time and training data are not required to be stored on servers. The challenge of learning feature representation in incremental scenarios has to do with the inherent problem of catastrophic forgetting, namely the loss of previously learned knowledge when new knowledge is assimilated  \cite{MCCLOSKEY1989109,ratcliff1990connectionist}.
Methods for alleviating catastrophic forgetting has been largely developed in the classification setting,
in which catastrophic forgetting 
is typically observed by a clear reduction in classification accuracy  \cite{vijayan2021continual,delange2021continual,masana2020class,parisi2019continual,belouadah2021comprehensive}. 
The fundamental differences with respect to learning internal feature representation for visual search tasks are: (1) evaluation metrics do not use classification accuracy (2) visual search data have typically a finer granularity with respect to categorical data and (3) no classes are required to be specifically learned. 
These differences might suggest different origins of the two catastrophic forgetting phenomena. In this regard, some recent works provide some evidence showing the importance of the specific task when evaluating the catastrophic forgetting of the learned representations  \cite{davari2021probing,chen2020exploration,pu2021lifelong,chen2021feature}. In particular, the empirical evidence presented  in \cite{davari2021probing} suggests that feature forgetting is not as catastrophic as classification forgetting.
We argue that such evidence is relevant in visual search tasks and that it can be exploited with techniques that learn incrementally without storing past samples in a memory buffer 
\cite{li2016learning}. 

According to this, in this paper, we propose a new distillation method for the continual representation learning task, in which the search performance degradation caused by feature forgetting is jointly mitigated while learning discriminative features. This is achieved by aligning current and previous features of the same class, while simultaneously pushing away features of different classes. 
We follow the basic working principle of contrastive loss \cite{chen2020simple} used in self-supervised learning, to effectively leverage label information in a distillation-based training procedure in which we replace anchor features with the feature of the teacher model.

Our contributions can be summarized as follows:
\begin{enumerate}
    \item We address the problem of continual representation learning proposing a novel method that leverages label information in a contrastive distillation learning setup. We call our method Contrastive Supervised Distillation (CSD).
    \item Experimental results on different benchmark datasets show that our CSD training procedure achieves state-of-the-art performance. 
    \item
    Our results confirm that  feature forgetting in visual retrieval using fine-grained datasets is not as catastrophic as in classification. 
\end{enumerate}

\section{Related Works}
\noindent
\textbf{Continual Learning (CL).}
CL has been largely developed in the classification setting, where methods have been broadly categorized based on exemplar
\cite{icarl,lucir,bic,pernici2021class} and regularization \cite{ewc,rwalk,li2016learning,jung2016less}. 
Only recently, continual learning for feature representation is receiving increasing attention and few works pertinent to the regularization-based category has been proposed \cite{chen2020exploration,pu2021lifelong,chen2021feature}.
The work in \cite{chen2020exploration} 
proposed an unsupervised alignment loss between old and new feature distributions according to the Mean Maximum Discrepancy (MMD) distance \cite{Gretton2009}.
The work \cite{chen2021feature} uses both the previous model and estimated features to compute a semantic correlation between representations during multiple model updates. The estimated features are used to reproduce the behaviour of older models that are no more available. 
Finally, \cite{pu2021lifelong} addresses the problem of lifelong person re-identification in which the previously acquired knowledge is represented as similarity graphs and it is transferred on the current data through graphs convolutions.
While these methods use labels only to learn new tasks, our method leverages labels information to both learn incoming tasks and for distillation. 

Reducing feature forgetting with feature distillation is also related to the recent backward compatible representation learning in which newly learned models can be deployed without the need to re-index the existing gallery images \cite{shen2020towards,pernici2021regular,biondi2021cores}. This may have an impact on privacy as also the gallery images are not required to be stored on servers. Finally, the absence of the cost re-indexing is advantageous in streaming learning scenarios as \cite{aljundi2019task,pernici2020self}.

\noindent
\textbf{Contrastive Learning} 
Contrastive learning has been proposed in~\cite{chopra2005learning} for metric learning and then it is demonstrated to be effective in unsupervised/self-supervised representation learning~\cite{he2020momentum,misra2020self,chen2020simple}.
All these works focus on obtaining discriminative representations that can be transferred to downstream tasks by fine-tuning.
In particular, this is achieved as, in the feature space, each image and its augmented samples (the positive samples) are grouped together {while} the others (the negative samples) are pushed away. 
However, \cite{KhoslaTWSTIMLK20} observed that, given an input image, samples of the same class are considered as negative and, consequently, pushed apart from it. 
We follow a similar argument which considers as positive also these images. 

\section{Problem Statement}

In the continual representation learning problem, a model $\mathrm{M}(\, \cdot \, ; \theta, \mathbf{W})$ is sequentially trained for $T$ tasks on a dataset $\mathcal{D} = \{ (\mathbf{x}_i, y_i, t_i) \, | \, i=1,2,\ldots,N \}$, where $\mathbf{x}_i$ is an image of a class $y_i \in \{1, 2, \ldots, L \}$, $N$ is the number of images, and $t_i \in \{ 1,2,\ldots,T \} \, $ is the task index associated to each image.
In particular, for each task $k$, $M$ is trained on the subset \mbox{$\mathcal{T}_k = \mathcal{D} |_{t_i = k} = \{ (\mathbf{x}_i, y_i, t_i) \, | \,  t_i = k \}$} which represents the $k$-th training-set that is composed by $L_k$ classes.
Each training-set {has different classes and images with respect to} the others and {only} $\mathcal{T}_k$ is available to train the model $\mathrm{M}$ (memory-free). 

At training time of task $k$, in response to a mini-batch $\mathcal{B} = \{ (\mathbf{x}_i, y_i, t_i) \}_{i=1}^{|\mathcal{B}|}$ of $\mathcal{T}_k$, the model $\mathrm{M}$ extracts the feature vectors and output logits for each image in the batch, i.e.,  $\mathrm{M}(\mathbf{x}_i) = C(\phi(\mathbf{x}_i))$, where $\phi(\cdot, \theta)$ is the representation model which extracts the feature vector $f_i = \phi(\mathbf{x}_i)$ and $C$ is the classifier, which projects the feature vector $f_i$ in an output vector $z_i = C(f_i)$. 
At the end of the training {phase}, $\mathrm{M}$ is used to index a gallery-set $\mathcal{G} = \{(\mathbf{x}_g, y_g) \, | \, g = 1, 2, \ldots, N_g \}$ according to the extracted feature vectors $\{ (f_g, y_g) \}_{g=1}^{N_g}$. 

At test time, a query-set $\mathcal{Q} = \{\mathbf{x}_q \, | \, q = 1, 2, \ldots, N_q \}$ is processed by the representation model $\phi(\cdot, \theta)$ in order to obtain the set of feature vectors $\{f_q \}_{q=1}^{N_q}$. 
According to {cosine} distance function $d$, the nearest sample in the gallery-set $\mathcal{G}$ is retrieved for each query sample $f_q$, i.e., 
\begin{equation}
    f^* = \argmin{g = 1,2, \ldots, N_g} d(f_g, f_q),
\end{equation}

\section{Method}
\makeatletter
\newcommand{\vast}{\bBigg@{3.3}}
\newcommand{\Vast}{\bBigg@{5}}
\makeatother
\begin{figure}[t]
    \centering
    \includegraphics[width=0.7\linewidth]{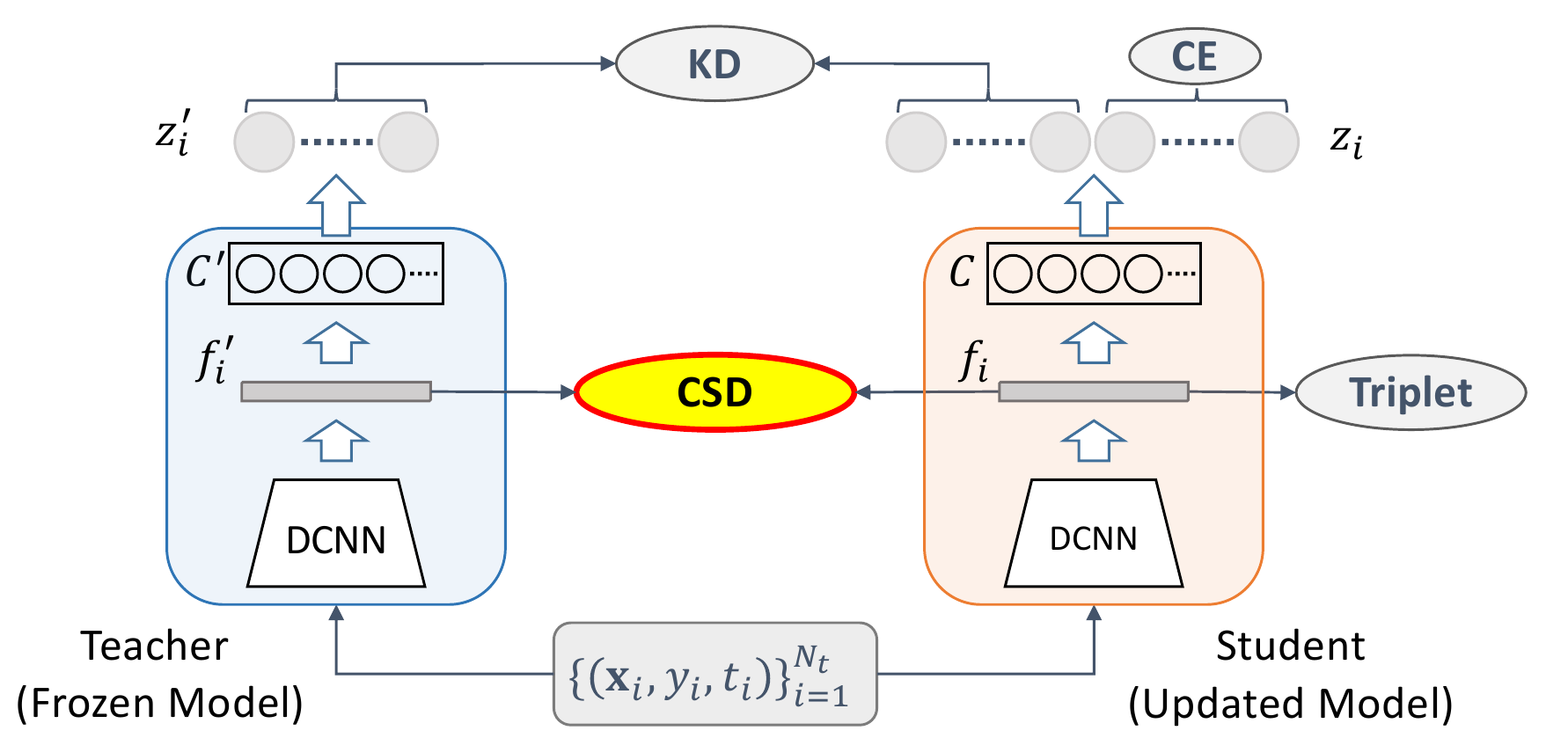}
    
    \vspace{-5pt}
    \caption{Proposed method is based on the teacher-student framework. During the training of the student, CE and triplet losses are minimized to learn the new task data, are KD and CSD are used to preserve the old knowledge using the teacher (not trainable).}
    \label{fig:teacher-student}
\end{figure}
To mitigate the effect of catastrophic forgetting while acquiring novel knowledge from incoming data, we propose a training procedure that follows the \textit{teacher-student} framework, {where the teacher is the model before the update and the student is the model that is updated. The teacher is leveraged during the training of the student to preserve the old knowledge as old data is not available.}

With reference to Fig.~\ref{fig:teacher-student}, at each task $k$, the student is trained on the training-set $\mathcal{T}_k = \{ (\mathbf{x}_i, y_i, t_i)\, | \, t_i = k \}$ and the teacher is set as frozen, i.e., not undergoing learning. 
The loss function that is minimized during the training of the student is the following: 
\begin{equation}
    \label{eq:global_loss}
    \mathcal{L} = \mathcal{L}_{plasticity} + \mathcal{L}_{stability}
\end{equation}
where $\mathcal{L}_{stability} = 0$ during the training of the model on the first task. 
In the following, the components of the plasticity and stability loss are analyzed in detail. 
In particular, we adopt the following notation. Given a mini-batch $\mathcal{B}$ of training data, both the student and the teacher networks produce a set of feature vectors and classifier outputs in response to training images $\mathbf{x}_i \in \mathcal{B}$. We refer to as $\{f_i \}$, $\{z_i \}$ for the feature vectors and classifier outputs of the student, respectively, with $\{f_i' \}$, $\{z_i' \}$ for the teacher ones, and with $|\mathcal{B}|$ to the number of elements in the mini-batch.

\subsection{Plasticity Loss}
Following~\cite{chen2020exploration}, during the training of the updated model, the plasticity loss is defined as follows:
\begin{equation}
    \label{eq:plasticity_loss}
    \resizebox{0.3\hsize}{!}{$
    \mathcal{L}_{plasticity} = \mathcal{L}_{\rm CE} + \mathcal{L}_{\rm triplet}
    $}
\end{equation}
with

\begin{equation}
    \label{eq:ce}
    \resizebox{0.45\hsize}{!}{$
    \mathcal{L}_{\rm CE} = \frac{1}{|\mathcal{B}|} \sum\limits_{i = 1}^{|\mathcal{B}|} \;  y_i \; \text{log}  \vast( \frac{\text{exp} \big( z_i \big) }{\sum_{j = 1}^{|\mathcal{B}|}\text{exp} \big( z_j \big) } \vast)
    $}
\end{equation}

\begin{equation}
    \label{eq:triplet}
    \resizebox{0.45\hsize}{!}{$
    \mathcal{L}_{\rm triplet} = {\rm max} \big( \vert\vert f_i - f_p  \vert\vert_2^2 - \vert\vert f_i - f_n \vert\vert_2^2 \big).
    $}
\end{equation}

$\mathcal{L}_{\rm CE}$ and $\mathcal{L}_{\rm triplet}$ are the cross-entropy loss and the triplet loss, respectively.
The plasticity loss of Eq.~\ref{eq:plasticity_loss} is optimized during the training of the model and it is used in order to learn the novel tasks. 

\subsection{Stability Loss}
The stability loss preserves the previously acquired knowledge in order to limit the catastrophic forgetting effect, {that is typically performed using the teacher model for distillation}. 
The stability loss we propose is formulated as follows:
\begin{equation}    \label{eq:stability_loss}
\resizebox{0.4\hsize}{!}{$
    \mathcal{L}_{stability} = \lambda_{\rm KD} \; \mathcal{L}_{\rm KD} + \lambda_{\rm CSD} \; \mathcal{L}_{\rm CSD}
    $}
\end{equation}
where $\lambda_{\rm KD}$ and $\lambda_{\rm CSD}$ are two weights factors {that} balance the two loss components, namely Knowledge Distillation (KD) and the proposed Contrastive Supervised Distillation (CSD).
In our experimental results, we set both $\lambda_{\rm KD}$ and $\lambda_{\rm CSD}$ to 1.
An evaluation of different values is reported 
in the ablation studies of Sec.~\ref{sec:ablation}. 

\noindent
\textbf{Knowledge Distillation.}
KD~\cite{hinton2015distilling} minimizes the log-likelihood between the classifier outputs of the student and the soft labels produced by the teacher, instead of the ground-truth labels ($y_i$) used in the standard cross-entropy loss. This encourages the outputs of the updated model to approximate the outputs produced by the previous one. 
KD is defined as follows:
\begin{equation}
    \label{eq:kd}
\resizebox{0.6\hsize}{!}{$
    \mathcal{L}_{\rm KD} = \frac{1}{|\mathcal{B}|} \sum_{i = 1}^{|\mathcal{B}|} \; \frac{\text{exp} \big( z_i' \big) } {\sum_{j = 1}^{|\mathcal{B}|} \text{exp} \big( z_j' \big) } \;  \text{ log} \vast( \frac{\text{exp} \big( z_i \big) }{\sum_{j = 1}^{|\mathcal{B}|} \text{exp} \big( z_j \big) } \vast)
    $}
\end{equation}
\noindent

\noindent
\textbf{Contrastive Supervised Distillation.}
\begin{figure}[t]
    \centering
    \hspace{-20pt}
    \subfigure[]{ \label{fig:csd_img_a}
        \adjincludegraphics[width=0.3\linewidth,trim={ 0. 0. {.57\width} 0.}, clip]{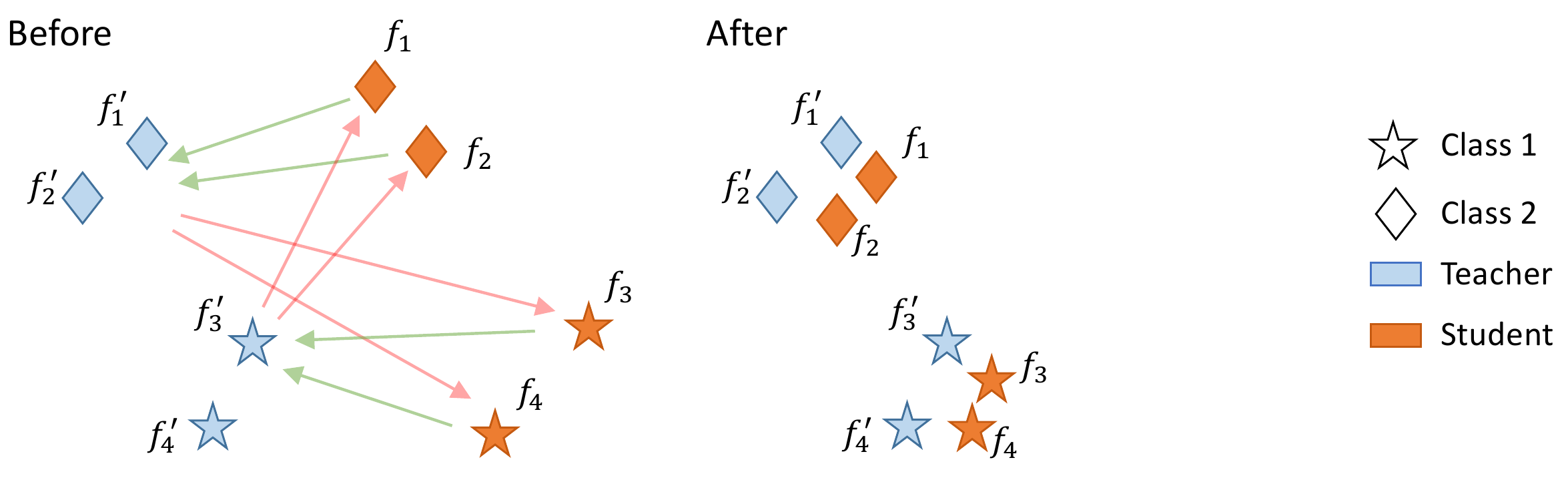}
    }
    \hspace{20pt}
    \subfigure[]{ \label{fig:csd_img_b}
        \adjincludegraphics[width=0.22\linewidth,trim={{.45\width} 0 {.2\width} 0.}, clip]{images/other/csd.pdf}
    }
    \subfigure{ \label{fig:csd_img_c}
        \adjincludegraphics[width=0.1\linewidth,trim={{.8\width} 0 0.  0.}, clip]{images/other/csd.pdf}
    }
    \vspace{-10pt}
    \caption{Proposed CSD loss. \textit{(a)} The features of four samples of two classes are firstly mapped in the feature space by the teacher (blue) and the student (orange). \textit{(b)} With CSD samples belonging to the same class (same symbol) are clustered together and separated from the others. }
    \label{fig:csd_img}
\end{figure}
We propose a new distillation loss, i.e., the Contrastive Supervised Distillation (CSD) that aligns current and previous feature models of the same classes while simultaneously pushing away features of different classes.
This is achieved at training time imposing the following loss penalty:
\begin{equation}
    \label{eq:csd}
    \resizebox{0.6\hsize}{!}{$
    \mathcal{L}_{\rm CSD} = - \frac{1}{|\mathcal{B}|} \sum\limits_{i = 1}^{|\mathcal{B}|} \frac{1}{|\mathcal{P}(i)|} \sum\limits_{p \in \mathcal{P}(i)}  \text{log} \vast( \frac{\text{exp} \big( f_i' \cdot f_p \big) }{\sum\limits_{\substack{{a = 1}\\{a \neq i}}}^{|\mathcal{B}|} \text{exp} \big( f_i' \cdot f_a \big) } \vast)
    $}
\end{equation}
where $\mathcal{P}(i) = \{(x_p, y_p, t_p) \in \mathcal{B} \, | \, y_p = y_i \}$ is a set of samples in the batch which belong to the same class of $\mathbf{x}_i$, i.e., the \textit{positive} samples.
Eq.~\ref{eq:csd} encourage for each class, the alignment of the student representations to the ones of the same class of the teacher model, which acts as anchors. In Fig.~\ref{fig:csd_img}, we show the effect of CSD loss on four samples $\{ (\mathbf{x}_i, y_i) \}_{i=1}^4$ with $y_i \in \{1, 2\}$. 
Initially (Fig.~\ref{fig:csd_img_a}) the feature vectors extracted by the student $f_i$ (orange samples) are separated from the teacher ones $f_i'$ (blue samples). CSD clusters together features of the same class moving the student representations, which are trainable, towards the fixed ones of the teacher {while pushing} apart features belonging to different classes. 
For the sake of simplicity, this effect is shown just for $f_1'$ and $f_3'$. Indeed, $f_1$ and $f_2$ become closer to $f_1'$, while $f_3$ and $f_4$ are spaced apart with respect to $f_1'$ as they are of class $2$. 
The same effect is visible also for $f_3'$ which attracts $f_3$ and $f_4$ and push away $f_1$ and $f_2$ as shown in Fig.~\ref{fig:csd_img_b}.

CSD imposes a penalty on feature samples considering not only the overall distribution of features of the teacher model with respect to the student one, but it also clusters together samples of the same class separating from the clusters of the other classes. 
Our method differs from KD as the loss function is computed directly on the features and not on the classifier outputs resulting in more discriminative representations. 
CSD also considers all the samples of each class as positive samples that are aligned with the same anchor of the teacher and not pairs (teacher-student) of samples as in \cite{romero2014fitnets}.  

\section{Experimental Results}
We perform our experimental evaluation on CIFAR-100~\cite{krizhevsky2009learning} and two fine-grained datasets, namely CUB-200~\cite{wah2011caltech} and Stanford Dogs~\cite{dogs}.
The CIFAR-100 dataset consists of $60000$ $32 \times 32$ images in $100$ classes.
The CUB-200 dataset contains $11788$ $224 \times 224$ images of $200$ bird species. 
Stanford Dogs includes over $22000$ $224 \times 224$ annotated images of dogs belonging to $120$ species.

The continual representation learning task is evaluated following two strategies.
In CIFAR-100, we evenly split the dataset into $T$ training-set where the model is trained sequentially, using the open-source Avalanche library~\cite{Lomonaco_2021_CVPR}. The experiments are evaluated with $T=2,5,10$. 
In CUB-200 and Stanford Dogs, following \cite{oh2016deep}\cite{wang2019multi}, we use half of the data to pre-train a model and split the remaining data into $T$ training-set. CUB-200 is evaluated with $T=1,4,10$ while Stanford Dogs with $T=1$. 

\noindent
\textbf{Implementation Details.}
We adopt ResNet32~\cite{he2016deep}\footnote{\href{https://github.com/arthurdouillard/incremental_learning.pytorch}{https://github.com/arthurdouillard/incremental\_learning.pytorch}} as representation model architecture on CIFAR-100 with 64-dimension feature space.
We trained the model for $800$ epochs for each task using Adam optimizer with a learning rate of $1 \cdot 10^{-3}$ for the initial task and $1 \cdot 10^{-5}$ for the others. Random crop and horizontal flip are used as image augmentation.
Following~\cite{chen2021feature}, we adopt pretrained Google Inception~\cite{szegedy2015going} as representation model architecture on CUB-200 and Stanford Dogs with 512-dimension feature space.
We trained the model for $2300$ epochs for each task using with Adam optimizer with a learning rate of $1 \cdot 10^{-5}$ for the convolutional layers and $1 \cdot 10^{-6}$ for the classifier.
Random crop and horizontal flip are used as image augmentation.
We adopt \textsc{Recall@K}\cite{jegou2010product}\cite{oh2016deep} as performance metric using each image in the test-set as query and the others as gallery. 

\begin{table}[t] 
 \small
 \caption{Evaluation on CIFAR-100 of CSD and compared methods.}
\label{tab:cifar2task} 
 \footnotesize
 \setlength{\tabcolsep}{14pt}
 \centering 
\resizebox{!}{1.8cm}{
  \begin{tabular}{l c c c} 
         \toprule
         \multirow{1}{*}{\textsc{Method}} & \shortstack{\textsc{Recall@1} \\ \textsc{ (1-50)}} & \shortstack{\textsc{Recall@1} \\ \textsc{(51-100)}} & \shortstack{\textsc{Recall@1} \\ Average}  \\

         \midrule
         \text{Initial model} & 67.6 & 21.7  & 44.7  \\
         \midrule
         \text{Fine-Tuning}  & 37.4 & \textbf{64.1}  & 50.8  \\
         \text{LwF}~\cite{li2016learning}    & 64.0 & 59.4  & 61.7  \\
         \text{MMD loss}~\cite{chen2020exploration}  & 61.8 & 60.9  & 61.4  \\
         CSD (Ours)  & \textbf{65.1} & 61.6   & \textbf{63.4}  \\
        
        \midrule \midrule
        \text{Joint Training}  & 70.5  &  71.9  & 71.2  \\
         \bottomrule 
     \end{tabular}
     }
 \end{table}
\subsection{Evaluation on CIFAR-100}
We compare our method on CIFAR-100 dataset with the Fine-Tuning baseline, LwF~\cite{li2016learning}, and \cite{chen2020exploration} denoted as MMD loss. As an upper bound reference, we report the Joint Training performance obtained using all the CIFAR-100 data to train the model. 

We report in Tab.~\ref{tab:cifar2task} the scores obtained with $T=2$. 
In the first row, we show the Initial Model results, i.e., the model trained on the first half of data from CIFAR-100. 
Our approach achieves the highest recall when evaluated on the initial task and the highest recall on the second task between methods trying to preserve old knowledge, being second only to Fine-Tuning that focuses only on learning new data. 
This results in our method achieving the highest average recall value with an improvement of $\mathtt{\sim}2\%$ \textsc{Recall@1} with respect to LwF and MMD loss and $10.4\%$ with respect to the Fine-Tuning baseline.
The gap between all the continual representation learning methods and Joint Training is significant ($\mathtt{\sim}8\%$). 
This underlines the challenges of CIFAR-100 in a continual learning scenario since there is a noticeable difference in the appearance between images of different classes causing a higher {feature forgetting}. 

\begin{figure}[t]
    \centering
    \hspace{-10pt}
    \subfigure[CIFAR-100 with $T=5$]{
        \label{fig:cifar_5task}
        \includegraphics[width=0.4\linewidth]{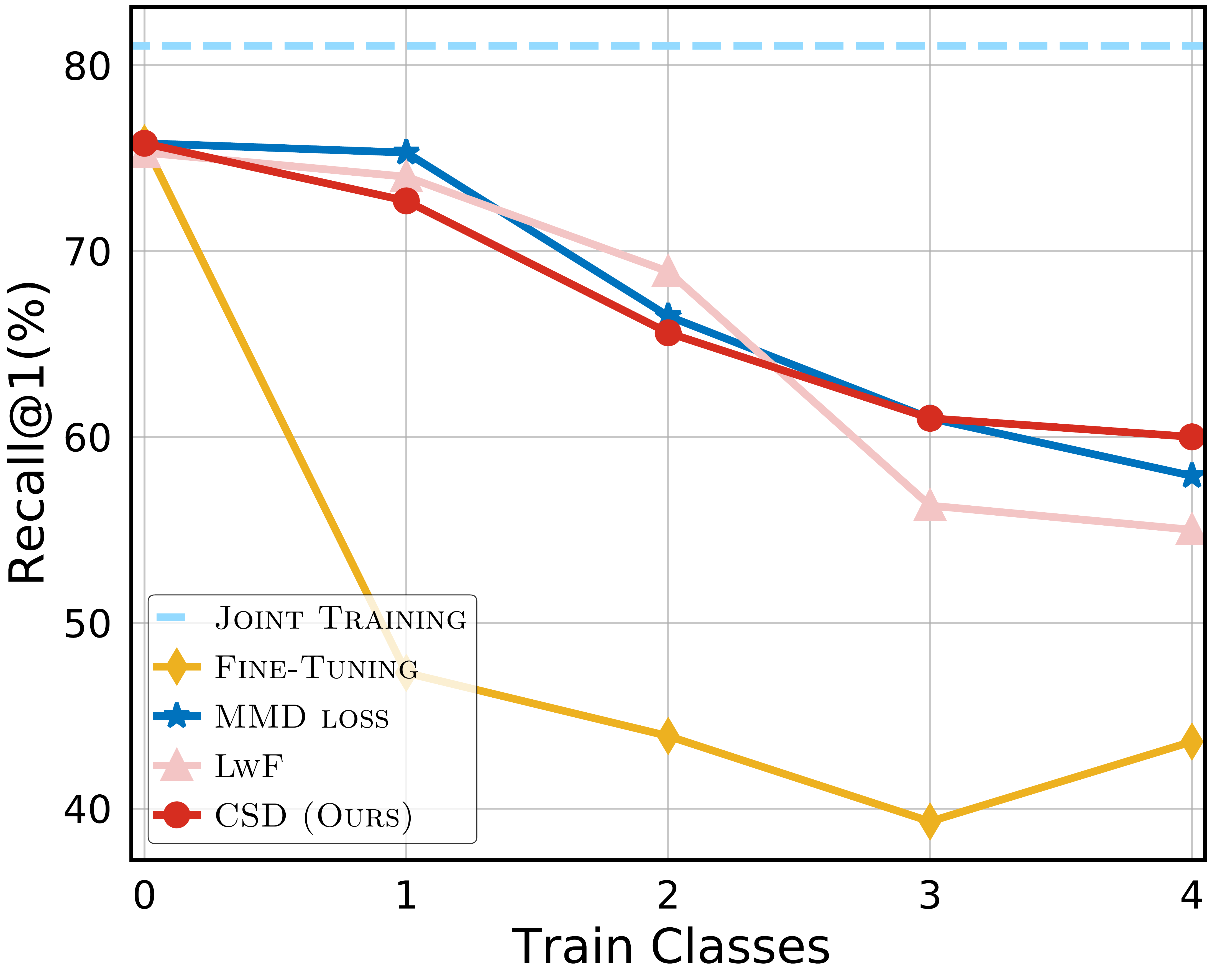}
    }
    \hspace{-5pt}
    \subfigure[CIFAR-100 with $T=10$]{
        \label{fig:cifar_10task}
        \includegraphics[width=0.4\linewidth]{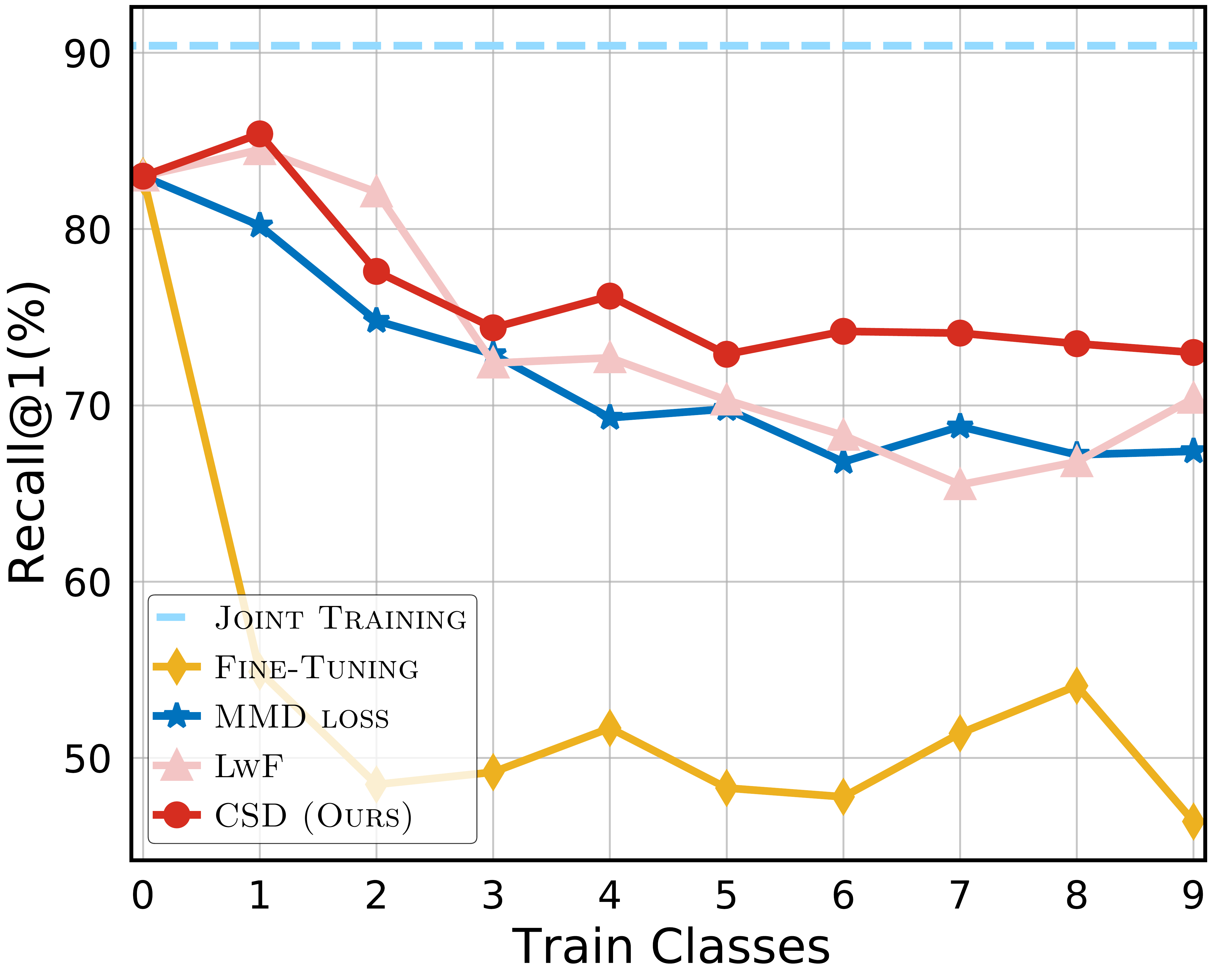}
    }
    
    \vspace{-10pt}
    \caption{Evolution of \textsc{Recall@1} on the first task as new tasks are learned on \mbox{CIFAR-100}. Comparison between our method (\textsc{CSD}) and compared methods.}
    \label{fig:cifar}
\end{figure}
Fig. \ref{fig:cifar_5task} and Fig. \ref{fig:cifar_10task} report the evolution of \textsc{Recall@1} on the initial task as new tasks are learned with $T=5$ and $T=10$, respectively.
In both experiments, our approach does not always report the highest scores, but it achieves the most stable trend obtaining the best result as the training end. 
This confirms that our approach is effective also when the model is updated multiple times.

\subsection{Evaluation on Fine-grained Datasets}

We compare our method on CUB-200 and Stanford Dogs datasets with the Fine-Tuning baseline, MMD loss~\cite{chen2020exploration}, and \cite{chen2021feature} denoted as Feature Estimation.
As an upper bound reference, we report the Joint Training performance obtained using all the data to train the model.

\begin{table}[t] 
\caption{Evaluation on Stanford Dogs and CUB-200 of CSD and compared methods.}
\label{tab:cub2task}
 \setlength{\tabcolsep}{3pt}
 \centering 
\resizebox{!}{1.9cm}{
  \begin{tabular}{l c c c c c c} 
         \toprule
         & \multicolumn{3}{c}{ \shortstack{\textsc{Stanford Dogs}} } & \multicolumn{3}{c}{ \shortstack{\textsc{CUB-200}} } \\
         
         \cmidrule(lr){2-4} 
         \cmidrule(lr){5-7}

         \shortstack{\textsc{Method}} 
         & \shortstack{\textsc{Recall@1} \\ \textsc{ (1-60)}} & \shortstack{\textsc{Recall@1} \\ \textsc{(61-120)}} & \shortstack{\textsc{Recall@1} \\ Average} 
         & \shortstack{\textsc{Recall@1} \\ (1-100)} & \shortstack{\textsc{Recall@1} \\  (101-200)} & \shortstack{\textsc{Recall@1} \\ Average} 
         \\

         \midrule
         \text{Initial model} & 81.3 & 69.3  & 75.3 & 79.2 & 46.9  & 63.1    \\
         \midrule
         \text{Fine-Tuning}   & 74.0 & \textbf{83.7} & 78.8 & 70.2 & 75.1 & 72.7\\
         \text{MMD loss} \cite{chen2020exploration} & 79.5  &  83.4 & 81.4  & 77.0 & 74.1  & 75.6   \\
         \text{Feat. Est.} \cite{chen2021feature}  & 79.9  &  83.5 & 81.7 & 77.7  &  75.0 & 76.4   \\
         CSD (Ours)  & \textbf{80.9} & 83.5 & \textbf{82.2}  & \textbf{78.6} & \textbf{78.3} & \textbf{78.5}  \\
        
        \midrule \midrule
         \text{Joint Training}  & 80.4 & 83.1 & 81.7 & 78.2 & 79.2 & 78.7 \\
         \bottomrule 
     \end{tabular}
     } 
 \end{table}
\begin{figure}[t]
    \centering
    \hspace{-10pt}
    \subfigure[CUB-200 with $T=4$]{
        \label{fig:cub_5task}
        \includegraphics[width=0.4\linewidth]{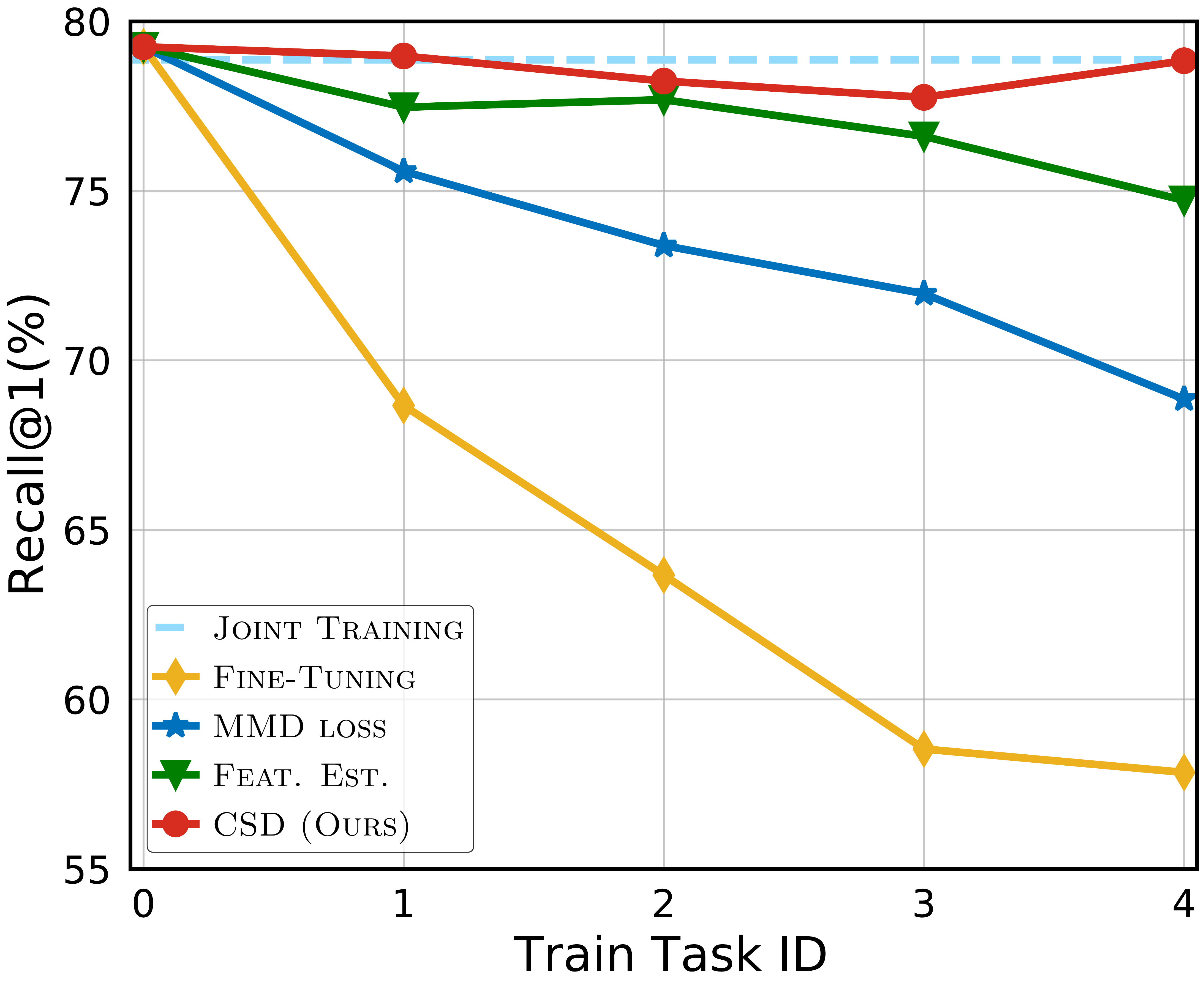}
    }
    \hspace{-5pt}
    \subfigure[CUB-200 with $T=10$]{
        \label{fig:cub_11task}
        \includegraphics[width=0.4\linewidth]{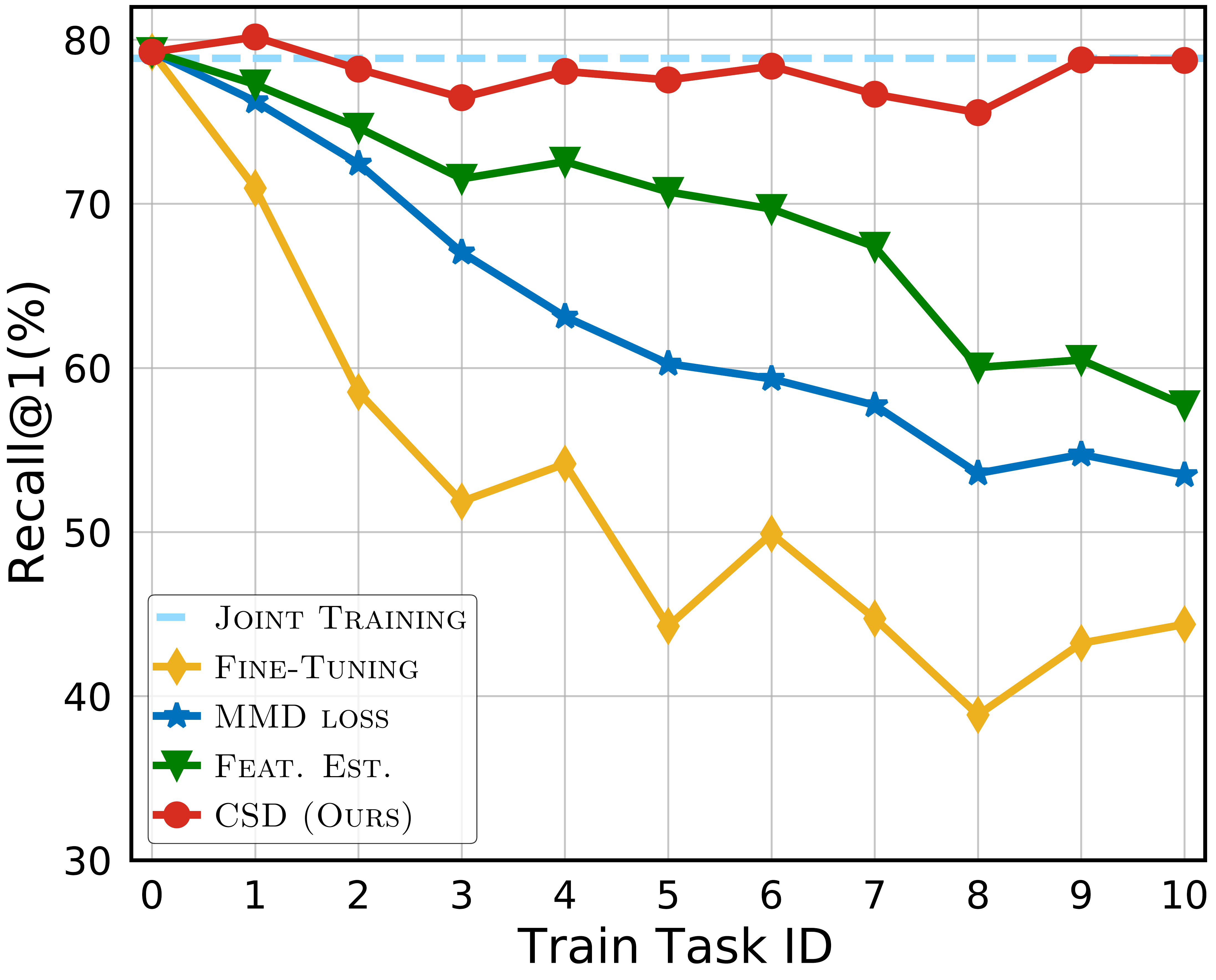}
    }
    \vspace{-10pt}
    \caption{Evolution of \textsc{Recall@1} on the first task as new tasks are learned on \mbox{CUB-200}. Comparison between our method (\textsc{CSD}) and compared methods.}
    \label{fig:cub}
\end{figure}
\begin{figure}[t]
\begin{minipage}{\textwidth}
  \begin{minipage}[b]{0.50\textwidth}
    \centering
    \includegraphics[width=0.88\linewidth]{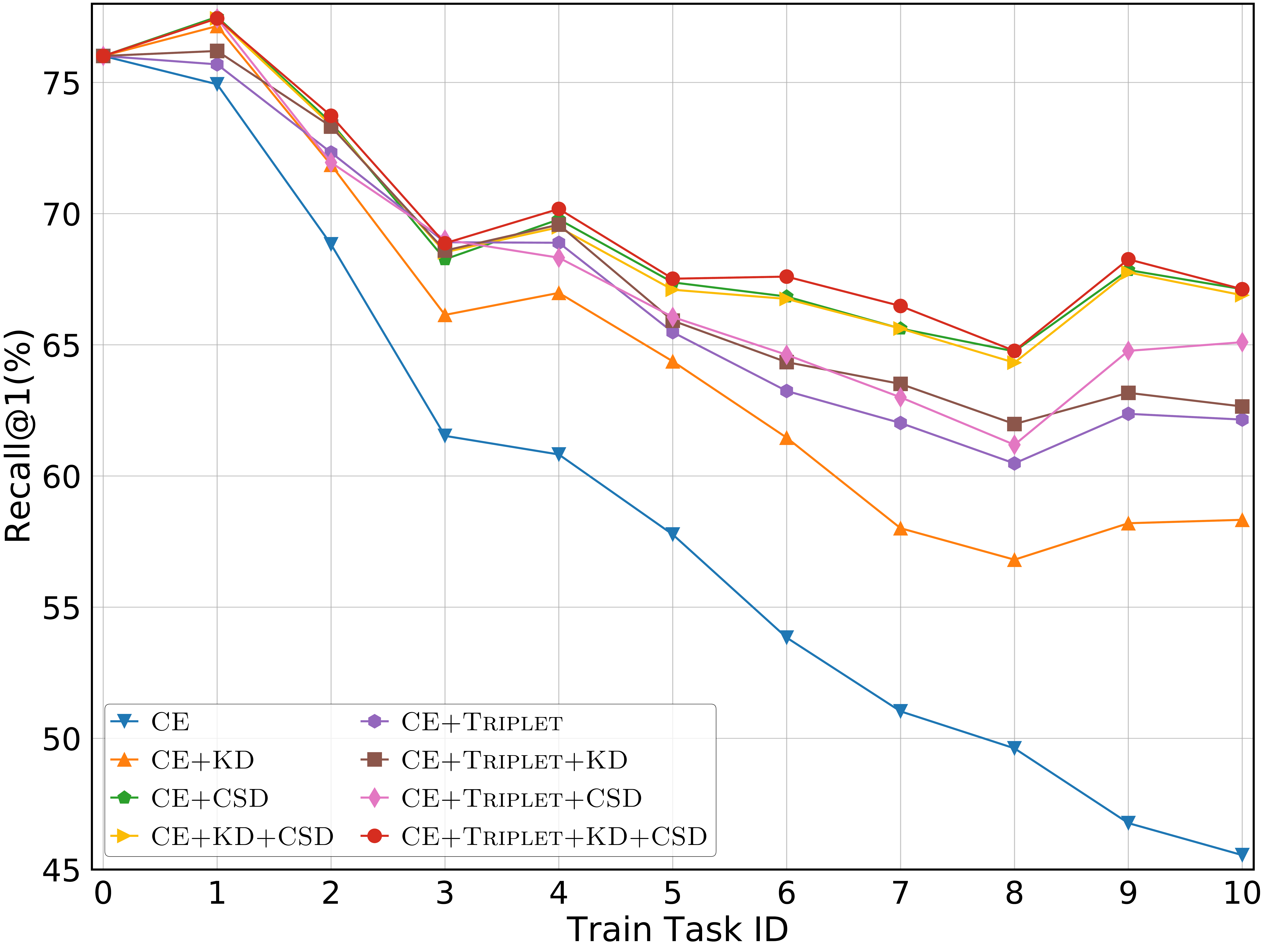}
    \vspace{-10pt}
    \captionof{figure}{Ablation on loss component on CUB-200  with $T=10$. ``+'' represents the combination of components.}
    \label{fig:abl_losscomp}
  \end{minipage}
  \hfill
  \begin{minipage}[b]{0.47\textwidth}
    \centering
    \resizebox{0.9\linewidth}{!}{
        \begin{tabular}{c c | c c c} 
            \toprule
            \multirow{1}{*}{$\lambda_{\rm KD}$} & \multirow{1}{*}{$\lambda_{\rm CSD}$} & \shortstack{\textsc{Recall@1} \\ \textsc{ (1-100)}} & \shortstack{\textsc{Recall@1} \\ \textsc{(101-200)}} & \shortstack{\textsc{Recall@1} \\ Average}  \\
            \midrule
            0.1 & 0.1 & 78.24 & 76.82  & 77.53  \\            
            0.1 & 1 & 79.19 & 77.50  & 78.35  \\              
            0.1 & 10 & 78.56 & 76.07  & 77.32  \\             
            \midrule
            1 & 0.1 & 79.32 & 73.82 & 76.57  \\              
            1 & 1  & 78.62  &  \textbf{78.34} & \textbf{78.48}  \\   
            1 & 10 & 79.12  & 75.32  &  77.22  \\            
            \midrule
            10 & 0.1 & 78.35 & 76.76 & 77.56 \\             
            10 & 1 & \textbf{79.53} & 76.93 & 78.23 \\              
            10 & 10 & 78.90 & 75.53 & 77.22 \\             
         
            \bottomrule 
            \hfill
        \end{tabular}
        }
        \vspace{-10pt}
        \captionof{table}{Ablation on the weight factors for KD and CSD in Eq.~\ref{eq:stability_loss} on CUB-200 with $T=1$.}
      \label{tab:loss_comp_weights}
    \end{minipage}
\end{minipage}
\end{figure}

We report in Tab.~\ref{tab:cub2task} the scores obtained with $T=1$ on the fine-grained datasets. 
On Stanford Dogs, our approach achieves the highest recall when evaluated on the initial task and comparable result with other methods on the final task with a gap of only $0.2\%$ with respect to Fine-Tuning that focus only on learning new data. 
This results in our method achieving the highest average recall value with an improvement of $0.5\%$ \textsc{Recall@1} concerning Feature Estimation, $0.8\%$ for MMD loss, and $3.4\%$ for Fine-Tuning.
On the more challenging CUB-200 dataset, we obtain the best \textsc{Recall@1} on both the initial and the final task outperforming the compared methods. 
{Our method achieves} the highest average recall value with an improvement of $2.1\%$ \textsc{Recall@1} with respect to Feature Estimation, $2.9\%$ for MMD loss, and $5.8\%$ for Fine-Tuning.
Differently from CIFAR-100, on fine-grained datasets, there is a lower dataset shift between different tasks leading to a higher performance closer to the Joint Training upper bound due to lower {feature forgetting}. 

We report in Fig.~\ref{fig:cub_5task} and Fig.~\ref{fig:cub_11task} the challenging cases of CUB-200 with $T=4$ and $T=10$, respectively.
These experiments show, consistently with Tab.~\ref{tab:cub2task}, how our approach outperforms state-of-the-art methods. 
In particular, with $T=10$ (Fig.~\ref{fig:cub_11task}), our method preserves the performance obtained on the initial task during every update. CSD largely improves over the state-of-the-art methods by almost $20\%$ - $25\%$ with respect to \cite{chen2021feature} and \cite{chen2020exploration} achieving similar performance to the Joint Training upper bound.  
By leveraging labels information for distillation during model updates, CSD
provides better performance and favorably mitigates the catastrophic forgetting of the representation compared to other methods that do not make use of this information.

\section{Ablation Study}\label{sec:ablation}

\noindent
\textbf{Loss Components.} In Fig.~\ref{fig:abl_losscomp}, we explore the benefits given by the components of the loss in Eq.~\ref{eq:global_loss} (i.e., CE, triplet, KD, and CSD) and their combinations in terms of \textsc{Recall@1} on CUB-200 with $T=10$. 
To observe single component performance, we analyze the trend of \textsc{Recall@1} on both the current task and previous ones evaluated jointly. When CSD is used, (i.e., CE+CSD, CE+KD+CSD, CE+triplet+CSD, CE+triplet+KD+CSD), we achieve higher \textsc{Recall@1} and maintain a more stable trend with respect to others. This underlines how CSD is effective and central to preserve knowledge and limit feature forgetting across model updates.  

\noindent
\textbf{Loss Components Weights.}
Finally, in Tab.~\ref{tab:loss_comp_weights}, we analyze the influence of the stability loss components varying the parameters $\lambda_{\rm KD}$ and $\lambda_{\rm CSD}$ of Eq.~\ref{eq:stability_loss} on CUB-200 with \mbox{$T=1$}. 
The table shows the \textsc{Recall@1} obtained on the first task, on the final task, and the average between them after training the model.
CSD best performs when  \mbox{$\lambda_{\rm KD} = \lambda_{\rm CSD} = 1$}, obtaining the highest average \textsc{Recall@1}. 

\section{Conclusions}

In this paper, we propose  Contrastive Supervised Distillation (CSD) to reduce feature forgetting in continual representation learning.
Our approach tackles the problem without storing data of previously learned tasks while learning a new incoming task.
CSD allows to minimize the discrepancy of new and old features belonging to the same class, while simultaneously pushing apart features from different classes of both current and old data in a contrastive manner.
We evaluate our approach and compare it to state-of-the-art works performing empirical experiments on three benchmark datasets, namely CIFAR-100, CUB-200, and Stanford Dogs. 
Results show the advantages provided by our method in particular on fine-grained datasets where CSD outperforms current state-of-the-art methods. 
Experiments also provide further evidence that  feature forgetting evaluated in visual retrieval tasks is not as catastrophic as in classification tasks.

\noindent
\textbf{Acknowledgments.}
This work was partially supported by the European Commission under European Horizon 2020 Programme, grant number 951911 - AI4Media.

\noindent
The  authors  acknowledge  the  CINECA award  under the ISCRA initiative (ISCRA-C - ``ILCoRe'', ID:~HP10CRMI87), for the availability of HPC resources.

\bibliographystyle{unsrt}
\bibliography{bib}

\begin{thebibliography}{10}

\bibitem{wan2014deep}
Ji~Wan, Dayong Wang, Steven Chu~Hong Hoi, Pengcheng Wu, Jianke Zhu, Yongdong
  Zhang, and Jintao Li.
\newblock Deep learning for content-based image retrieval: A comprehensive
  study.
\newblock In {\em Proceedings of the 22nd ACM international conference on
  Multimedia}, pages 157--166.

\bibitem{azizpour2014cnn}
H~Azizpour, J~Sullivan, S~Carlsson, et~al.
\newblock Cnn features off-the-shelf: An astounding baseline for recognition.
\newblock In {\em CVPRW}, pages 512--519. 2014.

\bibitem{yosinski2014transferable}
Jason Yosinski, Jeff Clune, Yoshua Bengio, and Hod Lipson.
\newblock How transferable are features in deep neural networks?
\newblock {\em Advances in Neural Information Processing Systems}, 2014.

\bibitem{chen2021deep}
Wei Chen, Yu~Liu, Weiping Wang, Erwin Bakker, Theodoros Georgiou, Paul Fieguth,
  Li~Liu, and Michael~S Lew.
\newblock Deep image retrieval: A survey.
\newblock {\em arXiv preprint arXiv:2101.11282}, 2021.

\bibitem{tolias2016particular}
Giorgos Tolias, Ronan Sicre, and Herv{\'e} J{\'e}gou.
\newblock Particular object retrieval with integral max-pooling of cnn
  activations.
\newblock In {\em ICLR 2016-International Conference on Learning
  Representations}, pages 1--12, 2016.

\bibitem{yue2015exploiting}
Joe Yue-Hei~Ng, Fan Yang, and Larry~S Davis.
\newblock Exploiting local features from deep networks for image retrieval.
\newblock In {\em Proceedings of the IEEE conference on computer vision and
  pattern recognition workshops}, pages 53--61, 2015.

\bibitem{price2019privacy}
W~Nicholson Price and I~Glenn Cohen.
\newblock Privacy in the age of medical big data.
\newblock {\em Nature medicine}, 25(1):37--43, 2019.

\bibitem{cossu2021sustainable}
Andrea Cossu, Marta Ziosi, and Vincenzo Lomonaco.
\newblock Sustainable artificial intelligence through continual learning.
\newblock {\em arXiv preprint arXiv:2111.09437}, 2021.

\bibitem{MCCLOSKEY1989109}
Michael McCloskey and Neal~J Cohen.
\newblock Catastrophic interference in connectionist networks: The sequential
  learning problem.
\newblock In {\em Psychology of learning and motivation}, volume~24, pages
  109--165. Elsevier, 1989.

\bibitem{ratcliff1990connectionist}
Roger Ratcliff.
\newblock Connectionist models of recognition memory: constraints imposed by
  learning and forgetting functions.
\newblock {\em Psychological review}, 97(2):285, 1990.

\bibitem{vijayan2021continual}
Mochitha Vijayan and SS~Sridhar.
\newblock Continual learning for classification problems: A survey.
\newblock In {\em International Conference on Computational Intelligence in
  Data Science}, pages 156--166. Springer, 2021.

\bibitem{delange2021continual}
Matthias Delange, Rahaf Aljundi, Marc Masana, Sarah Parisot, Xu~Jia, Ales
  Leonardis, Greg Slabaugh, and Tinne Tuytelaars.
\newblock A continual learning survey: Defying forgetting in classification
  tasks.
\newblock {\em IEEE Transactions on Pattern Analysis and Machine Intelligence},
  2021.

\bibitem{masana2020class}
Marc Masana, Xialei Liu, Bartlomiej Twardowski, Mikel Menta, Andrew~D Bagdanov,
  and Joost van~de Weijer.
\newblock Class-incremental learning: survey and performance evaluation on
  image classification.
\newblock {\em arXiv preprint arXiv:2010.15277}, 2020.

\bibitem{parisi2019continual}
German~I Parisi, Ronald Kemker, Jose~L Part, Christopher Kanan, and Stefan
  Wermter.
\newblock Continual lifelong learning with neural networks: A review.
\newblock {\em Neural Networks}, 113:54--71, 2019.

\bibitem{belouadah2021comprehensive}
Eden Belouadah, Adrian Popescu, and Ioannis Kanellos.
\newblock A comprehensive study of class incremental learning algorithms for
  visual tasks.
\newblock {\em Neural Networks}, 135:38--54, 2021.

\bibitem{davari2021probing}
MohammadReza Davari and Eugene Belilovsky.
\newblock Probing representation forgetting in continual learning.
\newblock In {\em NeurIPS 2021 Workshop on Distribution Shifts: Connecting
  Methods and Applications}, 2021.

\bibitem{chen2020exploration}
Wei Chen, Yu~Liu, Weiping Wang, Tinne Tuytelaars, Erwin~M. Bakker, and
  Michael~S. Lew.
\newblock On the exploration of incremental learning for fine-grained image
  retrieval.
\newblock In {\em {BMVC}}. {BMVA} Press, 2020.

\bibitem{pu2021lifelong}
Nan Pu, Wei Chen, Yu~Liu, Erwin~M Bakker, and Michael~S Lew.
\newblock Lifelong person re-identification via adaptive knowledge
  accumulation.
\newblock In {\em Proceedings of the IEEE/CVF Conference on Computer Vision and
  Pattern Recognition}, pages 7901--7910, 2021.

\bibitem{chen2021feature}
Wei Chen, Yu~Liu, Nan Pu, Weiping Wang, Li~Liu, and Michael~S Lew.
\newblock Feature estimations based correlation distillation for incremental
  image retrieval.
\newblock {\em IEEE Transactions on Multimedia}, 2021.

\bibitem{li2016learning}
Zhizhong Li and Derek Hoiem.
\newblock Learning without forgetting.
\newblock {\em IEEE transactions on pattern analysis and machine intelligence},
  40(12):2935--2947, 2017.

\bibitem{chen2020simple}
Ting Chen, Simon Kornblith, Mohammad Norouzi, and Geoffrey Hinton.
\newblock A simple framework for contrastive learning of visual
  representations.
\newblock In {\em International conference on machine learning}, pages
  1597--1607. PMLR, 2020.

\bibitem{icarl}
Sylvestre{-}Alvise Rebuffi, Alexander Kolesnikov, Georg Sperl, and Christoph~H.
  Lampert.
\newblock icarl: Incremental classifier and representation learning.
\newblock In {\em {CVPR}}, pages 5533--5542. {IEEE} Computer Society, 2017.

\bibitem{lucir}
Saihui Hou, Xinyu Pan, Chen~Change Loy, Zilei Wang, and Dahua Lin.
\newblock Learning a unified classifier incrementally via rebalancing.
\newblock In {\em {CVPR}}, pages 831--839. Computer Vision Foundation / {IEEE},
  2019.

\bibitem{bic}
Yue Wu, Yinpeng Chen, Lijuan Wang, Yuancheng Ye, Zicheng Liu, Yandong Guo, and
  Yun Fu.
\newblock Large scale incremental learning.
\newblock In {\em {CVPR}}, pages 374--382. Computer Vision Foundation / {IEEE},
  2019.

\bibitem{pernici2021class}
Federico Pernici, Matteo Bruni, Claudio Baecchi, Francesco Turchini, and
  Alberto Del~Bimbo.
\newblock Class-incremental learning with pre-allocated fixed classifiers.
\newblock In {\em 2020 25th International Conference on Pattern Recognition
  (ICPR)}, pages 6259--6266. IEEE, 2021.

\bibitem{ewc}
James Kirkpatrick, Razvan Pascanu, Neil Rabinowitz, Joel Veness, Guillaume
  Desjardins, Andrei~A Rusu, Kieran Milan, John Quan, Tiago Ramalho, Agnieszka
  Grabska-Barwinska, et~al.
\newblock Overcoming catastrophic forgetting in neural networks.
\newblock {\em Proceedings of the national academy of sciences},
  114(13):3521--3526, 2017.

\bibitem{rwalk}
Gido~M Van~de Ven and Andreas~S Tolias.
\newblock Three scenarios for continual learning.
\newblock {\em arXiv preprint arXiv:1904.07734}, 2019.

\bibitem{jung2016less}
Heechul Jung, Jeongwoo Ju, Minju Jung, and Junmo Kim.
\newblock Less-forgetting learning in deep neural networks.
\newblock {\em arXiv preprint arXiv:1607.00122}, 2016.

\bibitem{Gretton2009}
A.~Gretton, AJ. Smola, J.~Huang, M.~Schmittfull, KM. Borgwardt, and
  B.~Sch{\"o}lkopf.
\newblock {\em Covariate shift and local learning by distribution matching}.
\newblock MIT Press, 2009.

\bibitem{shen2020towards}
Yantao Shen, Yuanjun Xiong, Wei Xia, and Stefano Soatto.
\newblock Towards backward-compatible representation learning.
\newblock In {\em Proceedings of the IEEE/CVF Conference on Computer Vision and
  Pattern Recognition}, pages 6368--6377, 2020.

\bibitem{pernici2021regular}
Federico Pernici, Matteo Bruni, Claudio Baecchi, and Alberto Del~Bimbo.
\newblock {R}egular {P}olytope {N}works.
\newblock {\em IEEE Transactions on Neural Networks and Learning Systems},
  2021.

\bibitem{biondi2021cores}
Niccolo Biondi, Federico Pernici, Matteo Bruni, and Alberto Del~Bimbo.
\newblock {C}o{R}e{S}: {C}ompatible {R}epresentations via {S}tationarity.
\newblock {\em arXiv preprint arXiv:2111.07632}, 2021.

\bibitem{aljundi2019task}
Rahaf Aljundi, Klaas Kelchtermans, and Tinne Tuytelaars.
\newblock Task-free continual learning.
\newblock In {\em Proceedings of the IEEE/CVF CVPR}, pages 11254--11263, 2019.

\bibitem{pernici2020self}
Federico Pernici, Matteo Bruni, and Alberto Del~Bimbo.
\newblock Self-supervised on-line cumulative learning from video streams.
\newblock {\em Computer Vision and Image Understanding}, 197:102983, 2020.

\bibitem{chopra2005learning}
Sumit Chopra, Raia Hadsell, and Yann LeCun.
\newblock Learning a similarity metric discriminatively, with application to
  face verification.
\newblock In {\em 2005 IEEE Computer Society Conference on Computer Vision and
  Pattern Recognition (CVPR'05)}, volume~1, pages 539--546. IEEE, 2005.

\bibitem{he2020momentum}
Kaiming He, Haoqi Fan, Yuxin Wu, Saining Xie, and Ross Girshick.
\newblock Momentum contrast for unsupervised visual representation learning.
\newblock In {\em Proceedings of the IEEE/CVF Conference on Computer Vision and
  Pattern Recognition}, pages 9729--9738, 2020.

\bibitem{misra2020self}
Ishan Misra and Laurens van~der Maaten.
\newblock Self-supervised learning of pretext-invariant representations.
\newblock In {\em Proceedings of the IEEE/CVF Conference on Computer Vision and
  Pattern Recognition}, pages 6707--6717, 2020.

\bibitem{KhoslaTWSTIMLK20}
Prannay Khosla, Piotr Teterwak, Chen Wang, Aaron Sarna, Yonglong Tian, Phillip
  Isola, Aaron Maschinot, Ce~Liu, and Dilip Krishnan.
\newblock Supervised contrastive learning.
\newblock In {\em NeurIPS}, 2020.

\bibitem{hinton2015distilling}
Geoffrey Hinton, Oriol Vinyals, and Jeff Dean.
\newblock Distilling the knowledge in a neural network.
\newblock {\em arXiv preprint arXiv:1503.02531}, 2015.

\bibitem{romero2014fitnets}
Adriana Romero, Nicolas Ballas, Samira~Ebrahimi Kahou, Antoine Chassang, Carlo
  Gatta, and Yoshua Bengio.
\newblock Fitnets: Hints for thin deep nets.
\newblock {\em arXiv preprint arXiv:1412.6550}, 2014.

\bibitem{krizhevsky2009learning}
Alex Krizhevsky, Geoffrey Hinton, et~al.
\newblock Learning multiple layers of features from tiny images.
\newblock 2009.

\bibitem{wah2011caltech}
Catherine Wah, Steve Branson, Peter Welinder, Pietro Perona, and Serge
  Belongie.
\newblock The caltech-ucsd birds-200-2011 dataset.
\newblock {\em Computation \& Neural Systems Technical Report}, 2011.

\bibitem{dogs}
Aditya Khosla, Nityananda Jayadevaprakash, Bangpeng Yao, and Li~Fei-Fei.
\newblock Novel dataset for fine-grained image categorization.
\newblock In {\em First Workshop on Fine-Grained Visual Categorization, IEEE
  Conference on Computer Vision and Pattern Recognition}, Colorado Springs, CO,
  June 2011.

\bibitem{Lomonaco_2021_CVPR}
Vincenzo Lomonaco, Lorenzo Pellegrini, Andrea Cossu, Antonio Carta, Gabriele
  Graffieti, Tyler~L. Hayes, Matthias De~Lange, Marc Masana, Jary Pomponi,
  Gido~M. van~de Ven, Martin Mundt, Qi~She, Keiland Cooper, Jeremy Forest, Eden
  Belouadah, Simone Calderara, German~I. Parisi, Fabio Cuzzolin, Andreas~S.
  Tolias, Simone Scardapane, Luca Antiga, Subutai Ahmad, Adrian Popescu,
  Christopher Kanan, Joost van~de Weijer, Tinne Tuytelaars, Davide Bacciu, and
  Davide Maltoni.
\newblock Avalanche: {A}n {E}nd-to-{E}nd {L}ibrary for {C}ontinual {L}earning.
\newblock In {\em Proceedings of the IEEE/CVF Conference on Computer Vision and
  Pattern Recognition (CVPR) Workshops}, pages 3600--3610, June 2021.

\bibitem{oh2016deep}
Hyun Oh~Song, Yu~Xiang, Stefanie Jegelka, and Silvio Savarese.
\newblock Deep metric learning via lifted structured feature embedding.
\newblock In {\em Proceedings of the IEEE conference on computer vision and
  pattern recognition}, pages 4004--4012, 2016.

\bibitem{wang2019multi}
Xun Wang, Xintong Han, Weilin Huang, Dengke Dong, and Matthew~R Scott.
\newblock Multi-similarity loss with general pair weighting for deep metric
  learning.
\newblock In {\em Proceedings of the IEEE/CVF Conference on Computer Vision and
  Pattern Recognition}, 2019.

\bibitem{he2016deep}
Kaiming He, Xiangyu Zhang, Shaoqing Ren, and Jian Sun.
\newblock Deep residual learning for image recognition.
\newblock In {\em Proceedings of the IEEE conference on computer vision and
  pattern recognition}, pages 770--778, 2016.

\bibitem{szegedy2015going}
Christian Szegedy, Wei Liu, Yangqing Jia, Pierre Sermanet, Scott Reed, Dragomir
  Anguelov, Dumitru Erhan, Vincent Vanhoucke, and Andrew Rabinovich.
\newblock Going deeper with convolutions.
\newblock In {\em Proceedings of the IEEE conference CVPR}, 2015.

\bibitem{jegou2010product}
Herve Jegou, Matthijs Douze, and Cordelia Schmid.
\newblock Product quantization for nearest neighbor search.
\newblock {\em IEEE transactions on pattern analysis and machine intelligence},
  33(1):117--128, 2010.

\end{thebibliography}

\end{document}